\documentclass{article}
\usepackage{ijcai19}

\usepackage{times}
\usepackage{soul}
\usepackage{url}
\usepackage[hidelinks]{hyperref}
\usepackage[utf8]{inputenc}
\usepackage[small]{caption}
\usepackage{graphicx}
\usepackage{amsmath}
\usepackage{booktabs}
\usepackage{algorithm}
\usepackage{algorithmic}
\usepackage{subfig}
\usepackage{lineno,hyperref}
\usepackage{amsfonts}
\usepackage{multirow}
\usepackage{bm}
\modulolinenumbers[5]
\usepackage{rotating}
\usepackage{booktabs}

\usepackage{amsmath}
\newtheorem{theorem}{Theorem}

\title{High Dimensional Bayesian Optimization via Supervised Dimension Reduction\footnote{Accepted by IJCAI2019}}

\author{
Miao Zhang$^{1,2}$\footnote{Contact Author}\and
Huiqi Li$^{1}$\footnote{Corresponding Author}\And
Steven Su$^{2}$\\
\affiliations
$^1$School of Information and Electronics, Beijing Institute of Technology, China\\
$^2$Faculty of Engineering and Information Technology, University of Technology Sydney, Australia\\
\emails
Miao.Zhang-2@student.uts.edu.au,huiqili@bit.edu.cn, Steven.Su@uts.edu.au
}

\begin{document}

\maketitle

\begin{abstract}

Bayesian optimization (BO) has been broadly applied to computational expensive problems, but it is still challenging to extend BO to high dimensions. Existing works are usually under strict assumption of an additive or a linear embedding structure for objective functions. This paper directly introduces a supervised dimension reduction method, Sliced Inverse Regression (SIR), to high dimensional Bayesian optimization, which could effectively learn the intrinsic sub-structure of objective function during the optimization. Furthermore, a kernel trick is developed to reduce computational complexity and learn nonlinear subset of the unknowing function when applying SIR to extremely high dimensional BO. We present several computational benefits and derive theoretical regret bounds of our algorithm. Extensive experiments on synthetic examples and two real applications demonstrate the superiority of our algorithms for high dimensional Bayesian optimization.
\end{abstract}

\section{Introduction}

Many machine learning tasks are zeroth-order optimization problems with expensive evaluation function, where the optimal points could be reached only by limited querying objective function. Bayesian Optimization (BO) has shown its superiority in noisy and expensive black-box optimization problems, which utilizes surrogate function (acquisition function) based on a computationally cheap probabilistic model to balance the exploration and exploitation during optimizing process, and Gaussian Process (GP) ~\cite{rasmussen-et-al:gp} is the most popular probabilistic model in Bayesian Optimization. The iteration of Bayesian Optimization is described as: 1) BO learns the probabilistic Gaussian Process model and construct the acquisition function based on queried points; 2) Maximize acquisition function to find the most promising point to be queried; 3) Evaluate $\textbf{x}^*_t$ and update queried points.

There are many successful applications of Bayesian Optimization in artificial intelligence community, including hyper-parameter tuning ~\cite{snoek2012practical,bergstra2011algorithms}, neural architecture search ~\cite{kandasamy2018neural}, robotics ~\cite{deisenroth2015gaussian} and more. Although BO works well in these applications with low-dimension or moderate-dimension, its applicability to high-dimensional problems is still challenging due to the statistical and computational challenges  ~\cite{wang2013bayesian}. Current theoretical research works suggest that Gaussian Process based Bayesian Optimization is exponentially difficult with dimension ~\cite{srinivas2012information}, as the theoretical efficiency of statistical estimation of function depends exponentially on dimension. Several research works have been proposed to tackle the two problems in high-dimensional Bayesian Optimization (HDBO) by assuming the function varies along an underlying low-dimensional subspace ~\cite{chen2012joint,djolonga2013high,wang2013bayesian,ulmasov2016bayesian,li2017high}. Another popular method to handle high-dimensional BO is assuming an additive form of objective function ~\cite{kandasamy2015high,li2016high,wang2017batched,rolland2018high}.

In this paper, we introduce a supervised dimension reduction method (Sliced Inverse Regression, SIR) ~\cite{li1991sliced} to high-dimensional Bayesian optimization, which is used as projection matrix and could automatically learn the intrinsic structure of objective function during the optimization. Different with ~\cite{ulmasov2016bayesian}, which uses unsupervised dimension reduction method (PCA), our method takes the statistical relationship between the input and response into consideration and could handle the situation that the number of observations is less than dimension $N\ll D$. Furthermore, a kernel trick is developed to reduce the computational complexity of SIR when it is applied to handle extreme-large dimensional data, and also provide a nonlinear generalization to make SIR to learn a nonlinear function. The contributions of the paper are summarized in the following. 
\begin{itemize}
    \item Firstly, we formulize a novel SIR-BO algorithm which directly introduces a supervised dimension reduction method SIR to high-dimensional Bayesian Optimization to automatically learn the intrinsic structure of objective function during the optimization. 
    \item Secondly, the dimension reduction method in our Bayesian Optimization is extended to nonlinear setting through a kernel trick, which not only greatly reduces the complexity of SIR to make it possible to apply SIR to extremely-high dimensional Bayesian optimization, but also extracts nonlinear components simultaneously.
    \item Thirdly, we theoretically analyze the algorithm and its regret bounds, and extensively conduct experiments on synthetic data and real-world applications. Experimental results illustrate the superiority of our methods and imply that automatically learning the intrinsic low-dimensional structure of objective functions may be a more appropriate way for high dimensional Bayesian optimization than setting a strict prior assumption.
\end{itemize}

\section{Background}

\subsection{ Bayesian Optimization}

Bayesian optimization is aimed to maximize an unknown function $f: \mathcal{X} \rightarrow \mathbb{R}$ based on observations, that $\mathcal{X}\subseteq \mathbb{R}^D$. BO is usually applied to black-box optimization problems, where  there is no prior information about the exact function but only able to obtain noisy function evaluations $y=f(\textbf{x})+\epsilon$ by querying several exact points $\textbf{x}\in {\mathcal{X}}$. There are two key components for Bayesian Optimization: prior and acquisition function. In this paper, we investigate Gaussian Process as the prior ~\cite{rasmussen-et-al:gp}, that given $t$ pairs of observation points $\textbf{x}_{1:t}$ and corresponding evaluations $y_{1:t}$, we have a joint distribution:
\begin{equation} \label{[1]}
\textbf{f}(\textbf{x}_{1:t})\sim \boldsymbol{N}(\textbf{m}(\textbf{x}_{1:t}),\textbf{K}(\textbf{x}_{1:t},\textbf{x}_{1:t}))
\end{equation}
where $\textbf{K}(\textbf{x}_{1:t},\textbf{x}_{1:t})_{i,j}=\kappa(\textbf{x}_i,\textbf{x}_j)$ is the covariance matrix based on kernel function $\kappa$. Given a new observation point $\textbf{x}^*$ , the poster predictive distribution is calculated as:
\begin{equation} \label{[2]}
f(\textbf{x}^*)|\mathcal{D}_t,\textbf{x}^* \sim \boldsymbol{N}({\mu}(\textbf{x}^*),\sigma(\textbf{x}^*))
\end{equation}
where $\mathcal{D}_t=\{\textbf{x}_{1:t},y_{1:t}\}$, $\mu(\textbf{x}^*)=\kappa(\textbf{x}^*,\textbf{x}_{1:t})(\textbf{K}(\textbf{x}_{1:t},\textbf{x}_{1:t})+\theta_{c}^{2}I)^{-1}y_{i:t}$, $\sigma(\textbf{x}^*)=\kappa(\textbf{x}^*,\textbf{x}^*)-\kappa(\textbf{x}^*,\textbf{x}_{1:t})(\textbf {K}(\textbf{x}_{1:t},\textbf{x}_{1:t})+\theta_{c}^{2}I)^{-1}\kappa (\textbf{x}^*,\textbf{x}_{1:t}) ^{T}$, $\theta_{c}$ is a noise parameter.
$\theta_{c}$ and the hyperparameters of kernel $\kappa$ could be learned by maximizing the negative logarithm of the marginal likelihood function:
\begin{equation} \label{[3]}
\resizebox{.91\linewidth}{!}{$
\displaystyle
l=-\frac{1}{2}log(\textbf{det}(\textbf{K}+\theta_{c}^{2}I))-\frac{1}{2}y^{T}(\textbf{K}+\theta_{c}^{2}I)^{-1}y-\frac{t}{2}log2\pi
$}
\end{equation}

Acquisition function is a surrogate function based on the prior model which is used to determine the most promising sampling point for evaluation. In this paper we investigate Gaussian processes with Upper Confidence Bound (GP-UCB), which is defined as:
\begin{equation} \label{[4]}
\alpha _{\mathbf{UCB}}(\mathbf{x} )=\mu(\mathbf{x})+\sqrt{\beta }\sigma (\mathbf{x})
\end{equation}
with a tunable $\beta$ to balance exploitation against exploration.

\subsection{Sliced Inverse Regression}

There are many popular approaches for dimension reduction in machine learning community, including PCA, ISOMAP, LLE, and Laplacian Eigenmaps \cite{cunningham2015linear}, to find the manifold and intrinsic dimension of the input, but all of them are unsupervised without the consideration of the statistical relationship between the input $\textbf{x}$ and response $y$. In contrast, sliced inverse regression (SIR) tries to find an effective subspace based on inversely regressing input $\textbf{x}$ from response $y$, which has shown its superiority on dimension reduction in machine learning applications. SIR is based on the assumption of existing low-dimensional and effective subspace for a regression model:
\begin{equation} \label{[5]}
y=f(\beta'_1\textbf{x},...,\beta'_d\textbf{x},\epsilon), \quad \beta_k,\textbf{x}\in\mathbb{R}^{D}
\end{equation}
where $d\ll D$ and $\{\beta_1,...,\beta_d\}$ are orthogonal vectors forming a basis of the subspace, and $\epsilon$ is a noise. $\{\beta'_1\textbf{x},...,\beta'_d\textbf{x}\}$ is the dimension reduction projection operator from $D$-high-dimensional space to $d$-low-effective space with sufficient relevant information in \textbf{x} about $y$. SIR utilizes the inverse regression to find effective dimension reduction (e.d.r) subspace who considers the conditional expectation $\mathbb{E}(\textbf{x}|\textbf{y}=y)$ where $\textbf{y}$ varies as the opposite to classical regression setting $\mathbb{E}(\textbf{y}|\textbf{x}=x)$. Li~\shortcite{li1991sliced} has shown that the regression curve of $\mathbb{E}(\textbf{x}|\textbf{y}=y)$ hides in e.d.r subspace $B$, where the dimension reduction directions $\{\beta_1,...,\beta_L\}$ could be found by solving following eigen decomposition problem:
\begin{equation} \label{[6]}
\Gamma\beta =\lambda \Sigma \beta 
\end{equation}
where $\Gamma =\text{Cov}(E(\textbf{x}|\textbf{y}))$ is between-slice covariance matrix and $\Sigma$ is the covariance matrix of $\textbf{x}$.

\section{Supervised dimension reduction based Bayesian optimization}

The supervised dimension reduction based Bayesian optimization usually contains two main stages ~\cite{djolonga2013high}: \textbf{subspace learning}, which is to find e.d.r directions to project inputs into low-dimension data; and executing \textbf{normal Bayesian optimization} on the learned subspace. We first utilize SIR to learn towards the actual low-dimensional subspace $A$ that: $f(\textbf{x})=g(A\textbf{x})$. Defining $\hat A$ as the learned approximated dimension reduction matrix that: $\hat f(\textbf{x})=\hat g(\hat{A}\textbf{x})=\hat g(\textbf{u})$, we optimize $\hat g(\textbf{u})$ based on BO in the effective subspace and update the approximate projection matrix $\hat{A}$ based on SIR in certain iterations to make it aligned closer with the actual matrix $A$. 

\subsection{Subspace learning and computational efficiency}

We learn the projection matrix based on SIR by solving the eigen decomposition problem:
\begin{equation} \label{[7]}
\hat\Gamma \beta=\lambda\hat\Sigma\beta
\end{equation}
and the first $d$ eigenvectors are composed as the approximate projection matrix. $\hat\Sigma$ and $\hat\Gamma$ are $D\times D$ matrix, and the computational complexity of solving  Eq.\eqref{[7]} is $\mathcal{O}(2D^3)$, that we need to calculate the inverse of covariance matrix $\Sigma^{-1}$ and then eigenvectors of $\Sigma^{-1}\hat\Gamma$. Although SIR has been applied to several dimension reduction applications, it is unrealistic to apply standard SIR to extremely-high dimensional Bayesian optimization, e.g. $D\geq 100000$. In this paper, we utilize two tricks to reduce the computation complexity when applying SIR to etremly-high dimensional Bayesian optimization: reduced SVD and kernelizing input which extends SIR into reproducing kernel Hilbert space (RKHS) in terms of kernels according to ~\cite{wu2008kernel}.

\subsubsection{Numerical method based on reduced SVD}

As suggested by ~\cite{yeh2009nonlinear}, the rank of slice covariance $\hat\Gamma$ is $(J-1)$, as:
\begin{equation} \label{[8]}
\hat\Gamma=WW^T
\end{equation}
where the $j$th column of $W$ is defined as $w_j=\sqrt{\frac{n_j}{n}}E(\textbf{x}_i|i\in S_j)$, and $J$ is number of slices in SIR. We set $UMU^T$ as the eigenvalue decomposition of $W^T\hat\Sigma^{-1}W$, and $U$ is $J\times (J-1)$, $M$ is $(J-1)\times (J-1)$ when we only consider eigenvectors with nonzero eigenvalues. Eq.\eqref{[7]} could be transformed as :
\begin{equation} \label{[9]}
\hat\Sigma^{-\frac{1}{2}}WW^T\hat\Sigma^{-\frac{1}{2}}z=\lambda z
\end{equation}
where $z=\hat\Sigma^{\frac{1}{2}}\beta$. We assume $ZM^\frac{1}{2}U^T$ is the SVD of $\hat\Sigma^{-\frac{1}{2}}W$, and each column of $Z=\hat\Sigma^{-\frac{1}{2}}WUM^{-\frac{1}{2}}$ is thus a solution of Eq.\eqref{[9]} when we set $\Lambda=M$, so that each column of
\begin{equation} \label{[10]}
V=\hat\Sigma^{-\frac{1}{2}}Z=\hat\Sigma^{-1}WUM^{-\frac{1}{2}}
\end{equation}
is a solution of Eq.\eqref{[7]}, and the e.d.r matrix can be constructed by the combination of the columns of $V$. In this way, the computational complexity to find an e.d.r direction contains: $\mathcal{O}(J^3)$ for the eigenvalue decomposition of $W^T\hat\Sigma^{-1}W$ and $\mathcal{O}(D^3)$ for the inverse of $\hat\Sigma$. Normally, we set $J=d+1$ in this paper, and all non-zero eigenvectors are composed as the dimension reduction matrix, where $d$ is the assumed low-dimension.

\begin{algorithm}[tb]
\caption{SIR-BO}
\label{alg:algorithm1}
\textbf{Input}: $n$, $T$, kernel $\kappa$
\begin{algorithmic}[1]
\STATE $D_1=\{\textbf{x}_{1:n},y_{1:n}\}\leftarrow $ randomly generate $n$ points and evaluations.\\
\STATE $\beta\leftarrow $ calculate dimension reduction directions based on Eq.\eqref{[10]}.
\FOR{$i=1,2,...$}
\STATE Built Gaussian process prior based on $\textbf{u}=\beta D$;
\STATE Find $\mathbf{x}^*\in \mathcal{X}$ with maximal acquisition function value: $\textbf{x}_t^*$: $\mathbf{x}^*=\textbf{argmax}_{\textbf{x}\in\mathcal{X}}\textbf{UCB}(\textbf{u}),\mathbf{u}=\beta\textbf{x}$;
\STATE $\mathcal{D}_{i+1}=\mathcal{D}_i\cup\{\mathbf{x}^*,y^*\}$;
\ENDFOR
\end{algorithmic}
\end{algorithm}

\begin{algorithm}[tb]
\caption{KISIR-BO}
\label{alg:algorithm2}
\textbf{Input}: $n$, $T$, kernel $\kappa$
\begin{algorithmic}[1]
\STATE  $D_1^\textbf{K}=\{\textbf{k}_{1:n},y_{1:n}\}\leftarrow $ kernelize randomly generated points $D_1$ through kernel trick.\\
\STATE  $V\leftarrow $ calculate e.d.r directions based on Eq.\eqref{[14]}.\\
\STATE  Transform the kernel data to subspace by $V$: $D_1^\textbf{K} \stackrel{V}{\longrightarrow}{D_V^\textbf{K}}_1$, and perform $\textbf{GP-UCB}$ on this subspace.
\FOR{$i=1,2,...$}
\STATE Built Gaussian process prior based on ${D_V^\textbf{K}}_i$;
\STATE Find $\mathbf{x}^*\in \mathcal{X}$ with maximal acquisition function value: $\textbf{x}_t^*$: $\mathbf{x}^*=\textbf{argmax}_{\textbf{x}\in\mathcal{X}}\textbf{UCB}(\textbf{u}),\mathbf{u}=V\textbf{k},\textbf{k}=\kappa (\mathbf{x},:)$;
\STATE $\mathcal{D}_{i+1}=\mathcal{D}_i\cup\{\mathbf{x}^*,y^*\}$, kernelize $D_{i+1}$ to get $D_{i+1}^\textbf{K}$, update $V$ and get ${D_V^\textbf{K}}_{i+1}$;
\STATE Update $\kappa$ for certain iterations;
\ENDFOR
\end{algorithmic}
\end{algorithm}

\subsubsection{SIR with Kernelized Input}

Even though we could efficiently reduce the computational complexity of SIR to $\mathcal{O}(J^3)+\mathcal{O}(D^3)$ by SVD, it is still impossible to directly apply SIR to extremely-high Bayesian optimization. Furthermore, SIR is only able to find the effective linear subspace, while hard to extract nonlinear features. The kernel trick is a popular way to provide a nonlinear transformation for algorithms, where the input data are represented by kernel function of pairwise comparisons ~\cite{wu2007regularized}. In this paper, we proposed Kernelized Input SIR (KISIR) to handle this situations, which utilizes kernel trick to extend SIR to nonlinear dimension reduction, and reduce the computational complexity greatly. 

\begin{figure*}
  \begin{minipage}{4.3cm}
      \includegraphics[width=4.8cm,height=3cm]{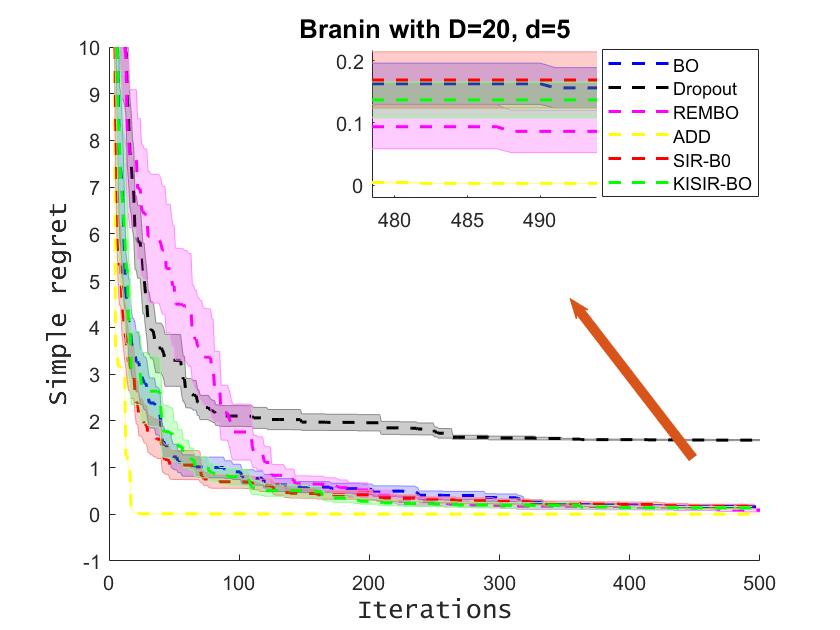}
  \end{minipage}
  \begin{minipage}{4.3cm}
      \includegraphics[width=4.8cm,height=3cm]{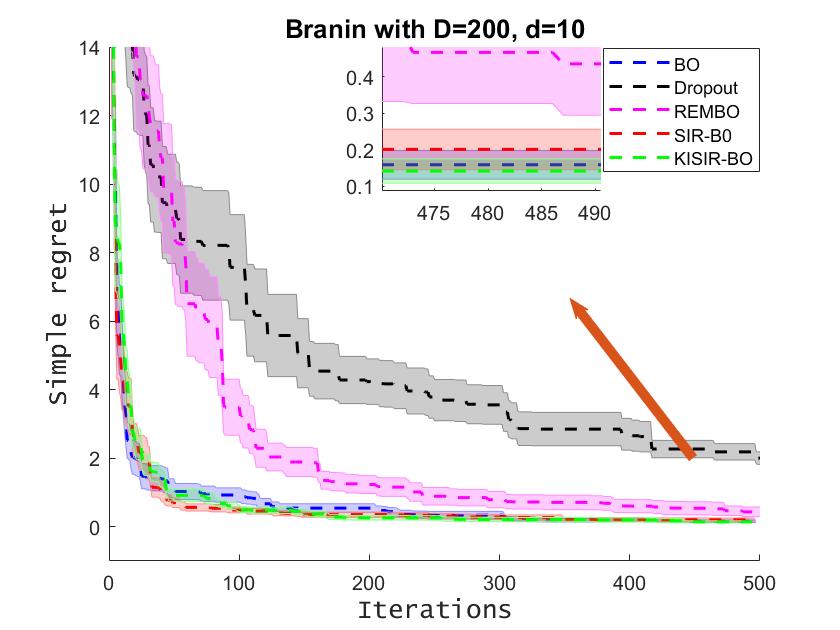}
  \end{minipage} 
  \begin{minipage}{4.3cm}
      \includegraphics[width=4.8cm,height=3cm]{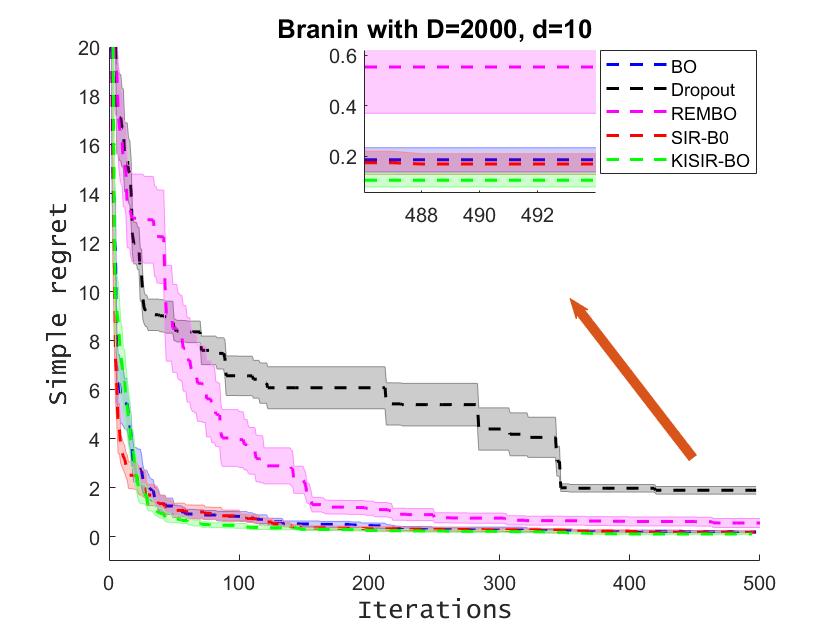}
  \end{minipage}
  \begin{minipage}{4.3cm}
      \includegraphics[width=4.8cm,height=3cm]{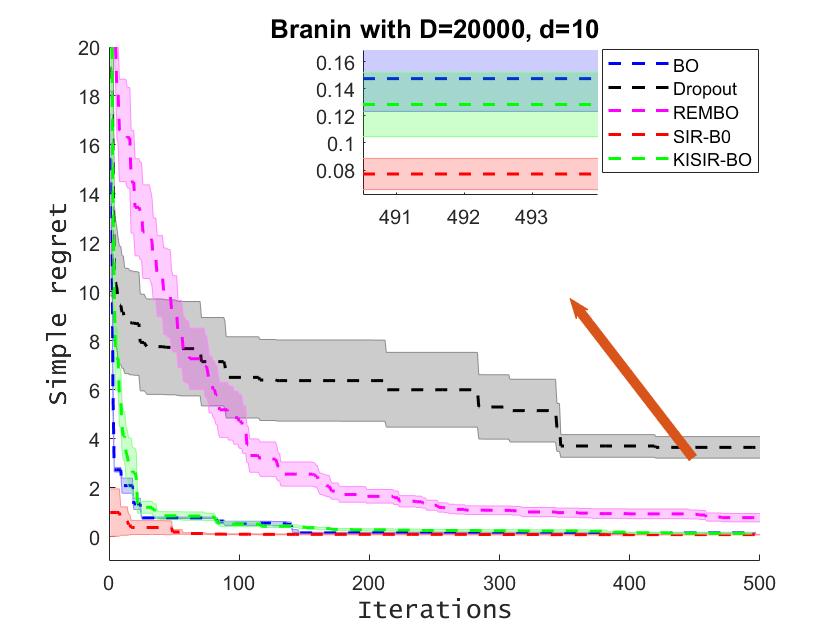}
  \end{minipage}
   \begin{minipage}{4.3cm}
      \includegraphics[width=4.8cm,height=3cm]{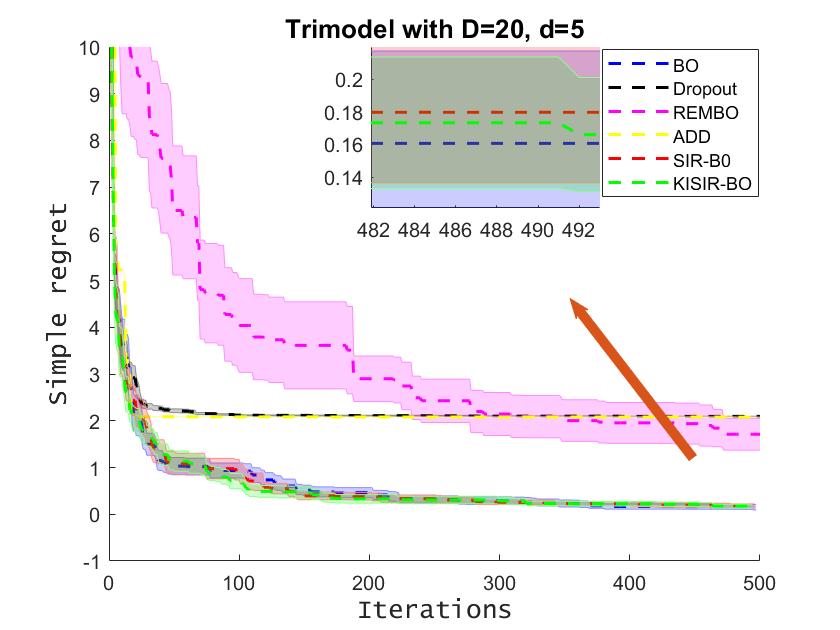}
  \end{minipage}
  \begin{minipage}{4.3cm}
      \includegraphics[width=4.8cm,height=3cm]{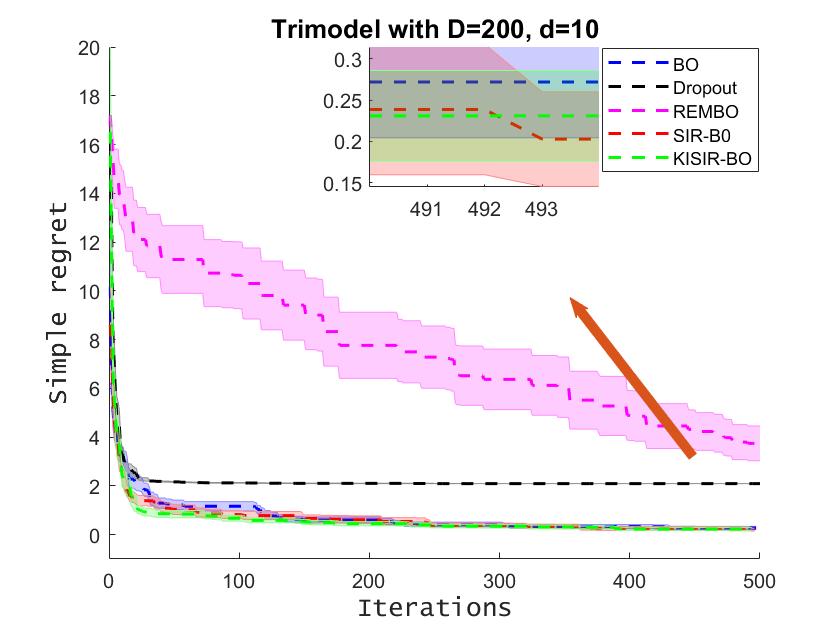}
  \end{minipage} 
  \begin{minipage}{4.3cm}
      \includegraphics[width=4.8cm,height=3cm]{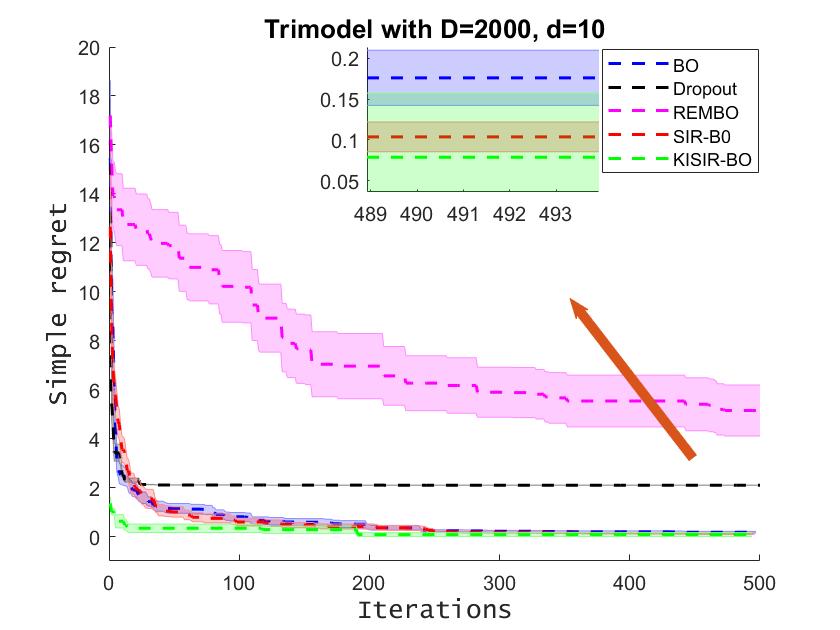}
  \end{minipage}
  \begin{minipage}{4.0cm}
      \includegraphics[width=4.8cm,height=3cm]{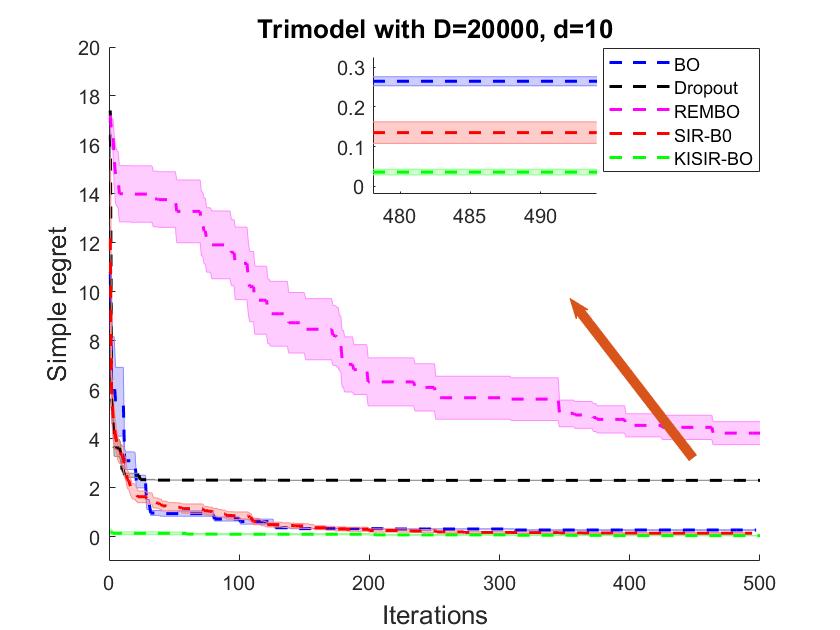}
  \end{minipage} 
  \caption{Simple regrets over iterations of our SIR-BO and KISIR-BO with compared approaches on two synthetic functions under different dimensions. We plot means with $1/4$ standard errors across 20 independent runs.}
 \label{figure1}
\end{figure*}

Given a kernel $\kappa$ and its spectrum: $\kappa(\textbf{x},\textbf{v})=\sum_{q=1}^{Q}\lambda_q \phi (\textbf{x}) \phi (\textbf{v}),\ Q\leq \infty ,\ \textbf{x},\textbf{v}\in \mathcal{X}$, the input space is transformed into spectrum-based feature space $\mathcal{H}_{\kappa}$: $\textbf{x} \mapsto \textbf{z}:\Phi (\textbf{x}):=(\sqrt{\lambda_1}\phi _1(\textbf{x}),\sqrt{\lambda_2}\phi _2(\textbf{x}),...)^T$, and $\left \langle \Phi (\textbf{x}),\Phi (\textbf{v}) \right \rangle _{\mathcal{H}_{\kappa}}=\kappa(\textbf{x},\textbf{v})$. So, the regression model with low-dimensional subspace as in Eq.\eqref{[5]} could be described as:
\begin{equation} \label{[11]}
y=f(\beta^{\mathcal{H}_\kappa}_1\textbf{z},...,\beta^{\mathcal{H}_\kappa}_d\textbf{z},\epsilon), \quad \beta^{\mathcal{H}_\kappa}_k \in\mathbb{R}^{Q},\ Q\leq \infty 
\end{equation}
in the feature space $\mathcal{H}_\kappa$. Consider $m_\textbf{z}=\sum_{i=1}^{n}\Phi (\textbf{x}_i)=0$ (otherwise center them according ~\cite{wu2008kernel}), we define $\hat\Sigma_{\textbf{z}}=\frac{1}{n}\sum_{i=1}^{n}\Phi (\textbf{x}_i)\Phi (\textbf{x}_i)^T$ as the sample covariance matrix of $\textbf{z}:=\Phi(\textbf{x})$, and $ \hat\Gamma_\textbf{z} =\sum_{j=1}^{J}\frac{n_j}{n}m_\textbf{z}^j m_\textbf{z}^{jT}$ as the sample between-slice covariance matrix, where $m_\textbf{z}^j=\frac{1}{n_j}\sum_{s\in S_j}\Phi(\textbf{x}_s)$ is the the mean for slice $S_j$ with $n_j$ size.  We could rephrase the eigenvalue decomposition of Eq.\eqref{[7]} in feature space $\mathcal{H}_\kappa$ as:
\begin{equation} \label{[12]}
\hat\Gamma_\textbf{z}\gamma =\lambda\hat\Sigma_\textbf{z}\gamma
\end{equation}
This problem could be solved by equivalent system:
\begin{equation} \label{[13]}
\left \langle  \Phi(\textbf{x}_i),\hat\Gamma_\textbf{z} \gamma \right \rangle_{\mathcal{H}_\kappa}=\lambda\left \langle \Phi(\textbf{x}_i),\hat\Sigma_\textbf{z}\gamma  \right \rangle_{\mathcal{H}_\kappa}
\end{equation}
where the solution is in the form of $\gamma=\sum_{i=1}^{N} \alpha _i\Phi(\textbf{x}_i)$. Based on previous definition, we could kernelize the input data as: $\textbf{K}=\{\kappa_{i,j}=\left \langle \Phi(\textbf{x}_i),\Phi(\textbf{x}_j) \right \rangle_{\mathcal{H}_k}=\kappa(\textbf{x}_i,\textbf{x}_j)\}_{N\times N}$, $\textbf{K}=\textbf{K}^T$. Define $\mathcal{H}_\kappa$ as the reproducing kernel Hilbert space based on $\kappa$, and the eigenvalue decomposition problem of SIR in $\mathcal{H}_k$ could be rephrased as:
\begin{equation} \label{[14]}
\hat\Gamma_\textbf{K}\gamma =\lambda\hat\Sigma_\textbf{K}\gamma
\end{equation}
where $\hat\Sigma_\textbf{K}=\textbf{K}\times \textbf{K}^T$, $\hat\Gamma_\textbf{K}=\sum_{j=1}^{J}m_\textbf{K}^j\times m_\textbf{K}^{jT}$, and $m_\textbf{K}^j=\frac{1}{n_j}\sum_{s\in S_j}\textbf{K}(:,s)$ is the mean for slice $S_j$ with $n_j$ size. The computational complexity of solving Eq.\eqref{[14]} is $\mathcal{O}(2N^3)$, where $N\ll D$ in extremely-high dimension reduction problem. Combining with the proof in Eq.\eqref{[10]}, the computational complexity could be reduced into $\mathcal{O}(J^3)+\mathcal{O}(N^3)$.

\subsection{Bayesian optimization in subspace}

The general process of SIR-BO is very similar to SI-BO in ~\cite{djolonga2013high}, and we also apply GP-UCB ~\cite{srinivas2012information} as the Bayesian optimizing method to theoretically analyze the regret bounds with ease, and perform it in an effective subspace. Algorithm~\ref{alg:algorithm1} presents a simple implementation of SIR-BO. The theoretical regret bounds of SIR-BO could also be derived based on Lemma 1, Theorem 4 and Theorem 5 in ~\cite{djolonga2013high}. However, after we kernelize input data through kernel trick, Bayesian optimization is supposed to be performed on a subspace of RKHS in our KISIR-BO, so we need to analyze transforming the input data into RKHS space trough kernel function $\kappa$: $\mathcal{X} \mapsto \mathcal{H}_\kappa$.

As defined in ~\cite{tan2018subspace}, we consider the function $f$ is randomly drawn from a distribution over the functions in the RKHS: $f=\sum_{i}\beta_i\Phi(\textbf{x}_i)$, where $\beta_i$ is Gaussian random variable, $\textbf{x}_i \in  \mathcal{X}:=\{\textbf{x}_1,\textbf{x}_2,...\}$ and the feature map $\Phi :\mathcal{X} \mapsto \mathcal{H}_\kappa$ is induced by kernel function $\kappa$. The function with kernerlized input $g(\textbf{k})$ is defined as:
\begin{equation} \label{[15]}
\resizebox{.85\linewidth}{!}{$
    \displaystyle
\begin{aligned}
& \{ g(\textbf{k}) =f(\textbf{x})|\textbf{x}\in \mathcal{X},\textbf{k}\in \mathcal{H}_\kappa, f=\sum _i \beta_i\Phi(\textbf{x}_i), \bm{\beta}\sim \mathcal{N}(\mathbf{b},\Sigma_{\beta})\}
\end{aligned}
$}
\end{equation}
where $\textbf{k}=\{\kappa(\textbf{x},\textbf{x}_1), \kappa(\textbf{x},\textbf{x}_2),...\}=\{\left \langle \Phi(\textbf{x}),\Phi(\textbf{x}_1) \right \rangle _{\mathcal{H}_\kappa},\left \langle \Phi(\textbf{x}),\Phi(\textbf{x}_2) \right \rangle _{\mathcal{H}_\kappa},...\}$, and the mean and covariance of $ g(\textbf{k})$ could be defined as:
\begin{equation} \label{[16]}
\resizebox{.85\linewidth}{!}{$
    \displaystyle
\begin{aligned}
m(g(\textbf{k})) & =m(f(\textbf{x}))=\sum _i b_i\left \langle \Phi(\textbf{x}),\Phi(\textbf{x}_i) \right \rangle_{\mathcal{H}_\kappa}\\
& =\sum _i b_i \kappa(\textbf{x},\textbf{x}_i)=\sum _i b_i \textbf{k}(i)\\
cov(g(\textbf{s}),g(\textbf{t})) & =\sum _i \sum _j (\Sigma_{\beta})_{ij}\left \langle \Phi(\textbf{s}),\Phi(\textbf{x}_i) \right \rangle_{\mathcal{H}_\kappa}\times  \\
& \left \langle \Phi(\textbf{t}),\Phi(\textbf{x}_i) \right \rangle_{\mathcal{H}_\kappa}
=\sum _i \sum _j (\Sigma_{\beta})_{ij}\textbf{s}(i)\textbf{t}(j)
\end{aligned}
$}
\end{equation}
$g(\textbf{k})$ are Gaussian random variables indexed by $\mathcal{X}$ which could be formulated as a Gaussian process~\cite{tan2018subspace}, and it is called as Kernelized Input Gaussian processes (KIGP) in this paper. The obvious difference between KIGP and GP is that the response difference between $\textbf{x}_i$ and $\textbf{x}_j$ is not a function of $\textbf{x}_i-\textbf{x}_j$ but of $\textbf{K}(\textbf{x}_i)-\textbf{K}(\textbf{x}_j)$, where $\textbf{K}(\textbf{z})=\{\kappa(\textbf{z},\textbf{x}_1),\kappa(\textbf{z},\textbf{x}_2),...\}$. It has been proved that KIGP is a strict superset of functions than GP with a same
trace-class covariance kernel function (see Theorem 1 in~\cite{tan2018subspace}).

\begin{figure*}
  \begin{minipage}{5.5cm}
      \includegraphics[width=5.5cm,height=3.5cm]{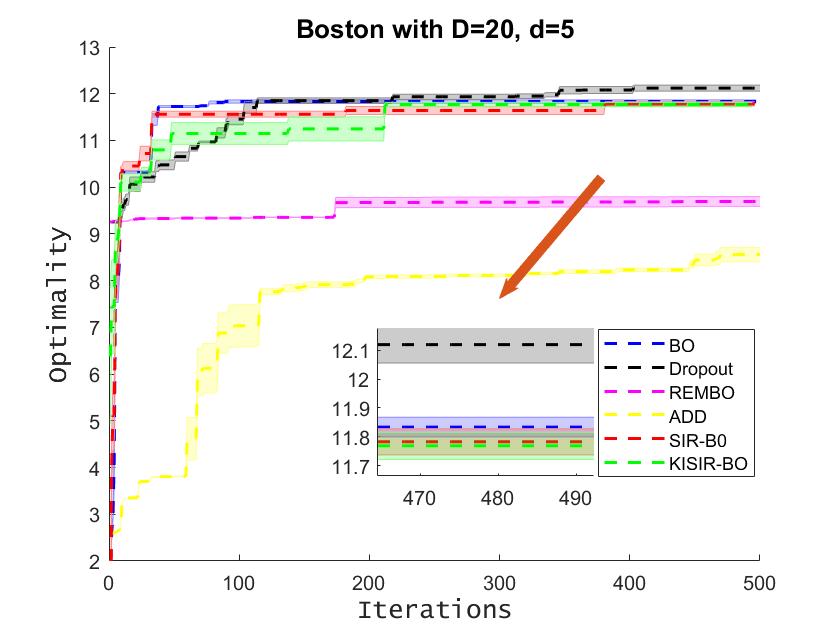}
  \end{minipage}
  \begin{minipage}{5.5cm}
      \includegraphics[width=5.5cm,height=3.5cm]{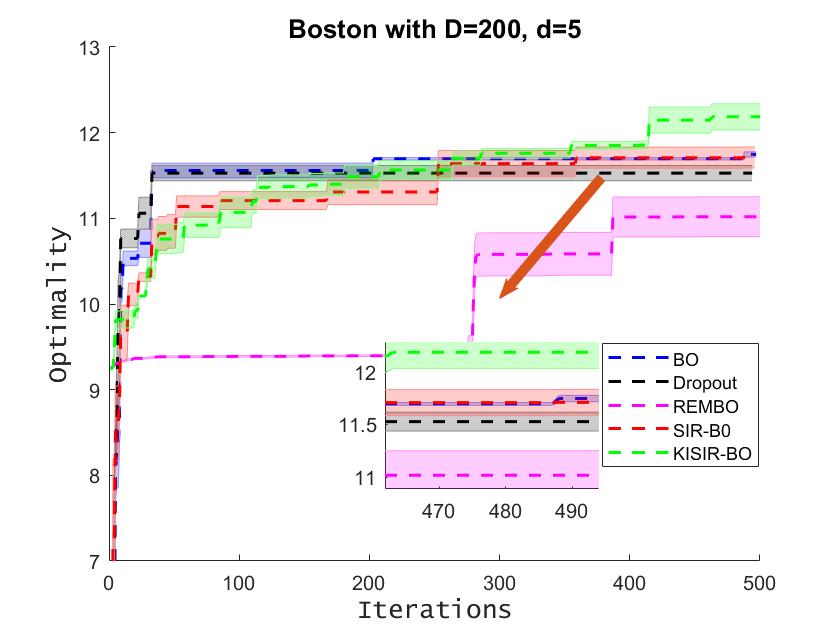}
  \end{minipage} 
  \centering
  \begin{minipage}{5.5cm}
      \includegraphics[width=5.5cm,height=3.5cm]{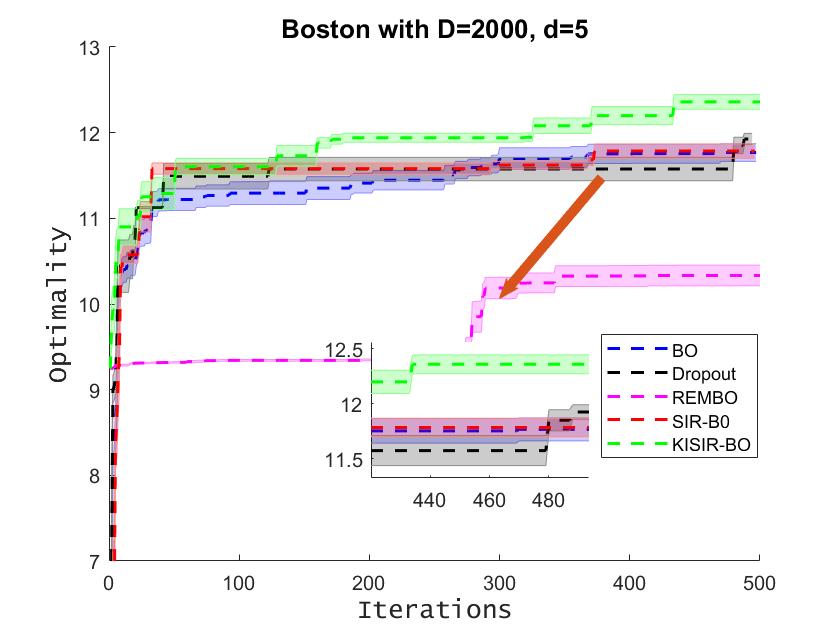}
  \end{minipage}
   \begin{minipage}{5.5cm}
      \includegraphics[width=5.5cm,height=3.5cm]{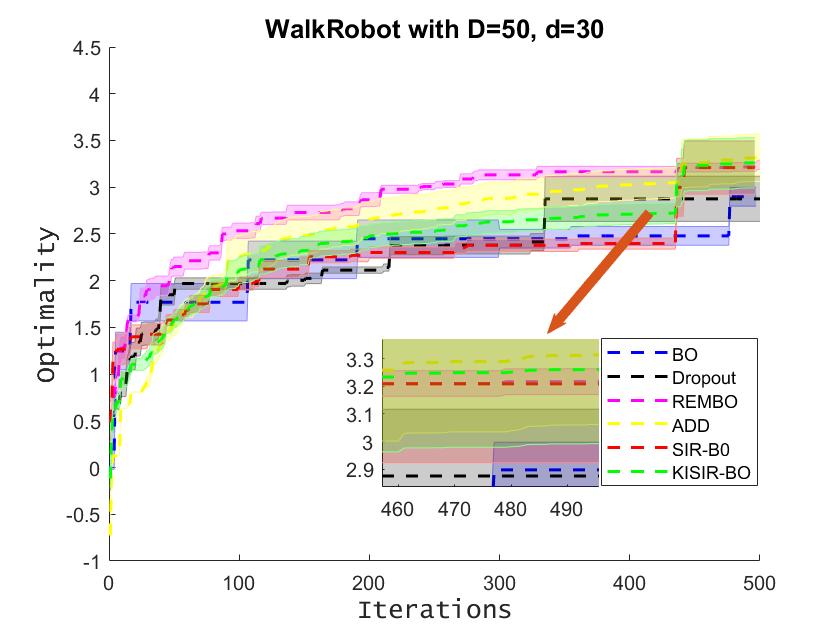}
  \end{minipage}
  \begin{minipage}{5.5cm}
      \includegraphics[width=5.5cm,height=3.5cm]{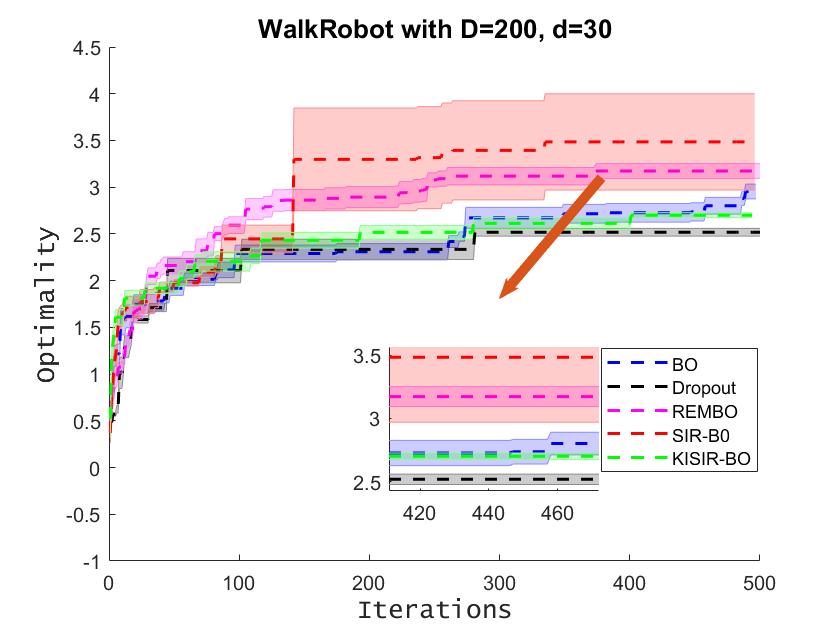}
  \end{minipage} 
  \centering
  \begin{minipage}{5.5cm}
      \includegraphics[width=5.5cm,height=3.5cm]{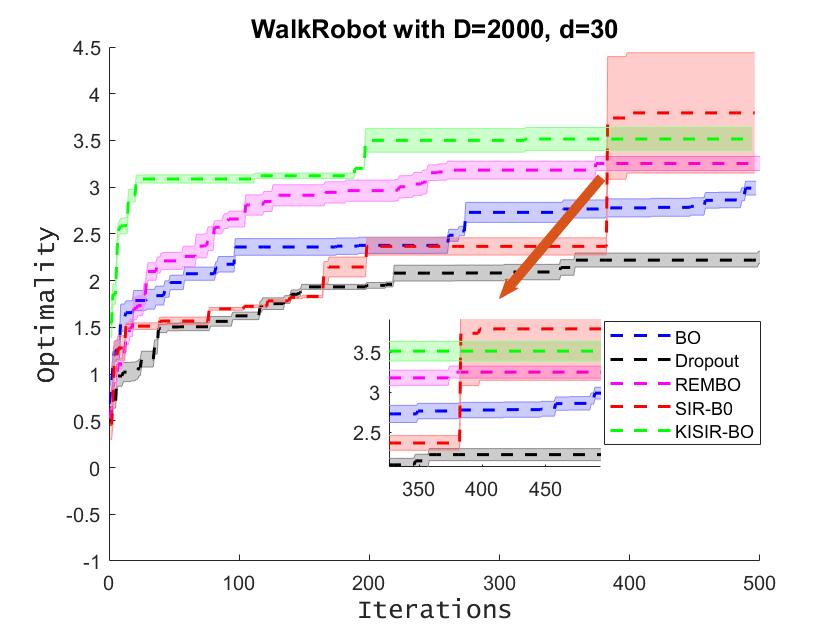}
  \end{minipage}
  \caption{Performance for configuring neural network on Boston dataset and controlling a three-link walk robot. We plot the mean and $1/4$ standard deviation for all comparing algorithms under different dimensions.}
 \label{figure2}
\end{figure*}

After proving $g(\textbf{k})$ is a Gaussian distribution and learning a dimension reduction matrix $V$ through SIR based on kernelized input data, we perform Bayesian optimization on an estimated subspace of $g(\textbf{k})$ : $\hat h(\textbf{u})$, which is also similar to ~\cite{djolonga2013high}. We give the cumulative regret bounds of KISIR-BO based on ~\cite{djolonga2013high} as follow:

\begin{theorem} Assume computational budget allows $T$ evaluations, and first spend $0< n\leq T$ sample evaluations to learn e.d.r. matrix $V$ which updates during the optimization. We also assume that with more evaluations, the learned e.d.r matrix $V$ is closer to actual matrix $A$, and $\left \| g-\hat g \right \|_\infty \leq \eta $, $\lVert \hat{h} \rVert \leq B $ where the error is bounded by $\hat \sigma$, each with probability at least $1-\delta/4$.

If perform $\textbf{GP-UCB}$ with $T-n$ steps on $\hat h(\textbf{u})$, with probability at least $1-\delta$, then the cumulative regret for KISRI-BO is bounded by:
\begin{equation} \label{[17]}
\resizebox{.85\linewidth}{!}{$
    \displaystyle
R_T\leq \underbrace {n}_{d.r.m\ learning}+\underbrace {(T-n)\eta}_{approx.\ error}+\underbrace {\mathcal{O}^*(\sqrt{TB}(\sqrt{\gamma_t}+\gamma_t))}_{R_{UCB}(T,\hat h,k)}
$}
\end{equation}
where $R_{UCB}(T,\hat h,\kappa)$ is the regret w.r.t. $\hat h$.
\end{theorem}

\newenvironment{proof}{\paragraph{Proof:}}{  $\square$}
\begin{proof}  As discussed above, $g(\textbf{k})$ could be formulated as a Gaussian process, and our KISIR-BO is also performed on the subspace of kernelized input space $\mathcal{H}_\kappa$. As defined in Lemma 1 in ~\cite{djolonga2013high}, the cumulative regret bounds for $\textbf{GP-UCB}$ performed on the subspace is $n+T\eta+\mathcal{O}^*(\sqrt{TB}(\sqrt{\gamma_t}+\gamma_t))$, and based on the assumption that, with more evaluations, the learned e.d.r matrix $V$ is closer to actual matrix $A$, so the $approx.\ error\leq (T-n)\eta$, and the regret bounds for our KISIR-BO could be obtained.
\end{proof}

Based on Theorem 3 in ~\cite{srinivas2012information} and Theorem 4 and 5 in ~\cite{djolonga2013high}, the regret bound could be rephrased as:
\begin{equation} \label{[18]}
\resizebox{.85\linewidth}{!}{$
    \displaystyle
\begin{aligned}
& R_T\leq \mathcal{O}(k^3d^2\mathbf{log}^2(1/\delta)) +2\mathcal{O}(\sqrt{TB}(\sqrt\gamma _t + \gamma _t))\\
& R_T\leq \mathcal{O}(\delta^2k^{11.5}d^7T^{4/5}\mathbf{log}^3(1/\delta)) +2\mathcal{O}(\sqrt{TB}(\sqrt\gamma _t + \gamma _t))
\end{aligned}
$}
\end{equation}
for noiseless observation and for noisy observations, respectively. ~\cite{srinivas2012information} gives the bounds $\gamma _t$ for different commonly used kernels.

Algorithm~\ref{alg:algorithm2} presents the details of KISIR-BO. We firstly need to define computational budget with $T$ evaluation and $n$ initial sample evaluations. $n$ randomly initial points with evaluation $D_1=\{\textbf{x}_{1:n},y_{1:n}\}$, which are preprocessed by kernel trick to get $D_1^\textbf{K}=\{\textbf{k}_{1:n},y_{1:n}\}$. We then calculate dimension reduction directions $V$ based on Eq.\eqref{[14]}, and transform the kernel data through $V$: $D_1^\textbf{K} \stackrel{V}{\longrightarrow}{D_V^\textbf{K}}_1$ and apply GP-UCB on the subspace ${D_V^\textbf{K}}_1$. The Bayesian optimization process is very similar to normal BO, while we need to notice that we could only infer $\textbf{u}$ from $\textbf{x}$ and it is impossible to recover $\textbf{x}$ from $\textbf{u}$ through $V$. Different from ~\cite{djolonga2013high} using a low-rank matrix recovery method to get an approximate $\hat{\textbf{x}}$, we perform heuristic algorithm (CMA-ES) directly on input space $\mathcal{X}$, and find the point $\textbf{x}_t^*$: $\mathbf{x}^*=\textbf{argmax}_{\textbf{x}\in\mathcal{X}}\textbf{UCB}(\textbf{u}),\mathbf{u}=V\textbf{k},\textbf{k}=\kappa (\mathbf{x},:)$. We update dimension reduction direction every iteration and update kernel hyperparameters for certain iterations.

\section{Experiments}

To evaluate the performance of our SIR-BO and KISIR-BO\footnote{Source code and used data are available at https://github.com/MiaoZhang0525}, we have conducted a set of experiments on two synthetic functions and two real datasets. We compare our approach against the standard BO, REMBO\footnote{The code of REMBO is available at \url{https://github.com/ziyuw/rembo} ~\cite{wang2013bayesian}.}, ADD-GP-UCB\footnote{The code of ADD-BO is available at \url{https://github.com/zi-w/Structural-Kernel-Learning-for-HDBBO}~\cite{kandasamy2015high}.}, and Dropout-BO\footnote{We implement Dropout-BO based on ~\cite{li2017high}.}. We adopt Gaussian kernel with lengthscale 0.1 for kernelizing input, and Gaussian kernel with adaptive lengthscale for Gaussian processes learned  by  maximizing marginal likelihood function through DIRECT, and optimize acquisition function by CMA-ES.

\subsection{Optimization on synthetic functions}

In this section, we demonstrate our approaches on two synthetic function: the first one is standard Branin function ~\cite{lizotte2008practical} with intrinsic dimension $d_e=2$, embedded into $D$-dimensional space, defined as:
\begin{equation} \label{[19]}
\resizebox{.85\linewidth}{!}{$
\begin{aligned}
    \displaystyle
f(\textbf{x})=g(\textbf{u})=a(x_j-b x_i^2+c x_i-r)+s(1-t)cos(x_i)+s
\end{aligned}
$}
\end{equation}
where $i$, $j$ are two randomly selected dimensions, $a=1$, $b=\frac{5.1}{4\pi^2}$, $c=\frac{5}{\pi}$, $r-6$, $s=10$, $t=\frac{1}{8\pi}$, and the global minimum is $g(\textbf{u}^*)=0.397887$ at $\textbf{u}^*=(-\pi,12,275),(\pi,2.275)$ and $(9.42478,2.475)$.  The second synthetic function is a trimodal function with $d_e=2$:
\begin{equation} \label{[20]}
\begin{aligned}
& f(\textbf{x})=g(\textbf{u})=\textup{log}(0.1\times\textup{mvnpdf}(\textbf{u},\textbf{c}_1,\sigma ^2)+\\
& 0.8\times\textup{mvnpdf}(\textbf{u},\textbf{c}_2,\sigma ^2)+0.1\times \textup{mvnpdf}(\textbf{u},\textbf{c}_3,\sigma ^2))
\end{aligned}
\end{equation}
where $\sigma^2=0.01d_e^{0.1}$ and $\textup{mvnpdf}(\textbf{x},\mu,\sigma^2)$ is multivariates Gaussian distribution and $c_i$ are fixed centers in $\mathbb{R}^{d_e}$. The global maximum is $f(\textbf{x}^*)=g(\textbf{u}^*)=g(\textbf{c}_2)=2.4748$ at the center $\textbf{c}_2$ with highest probability 0.8.

\begin{table}\tiny
\centering
\setlength{\tabcolsep}{6pt}
\begin{tabular}
{p{26pt}{l}p{21pt}{l}p{20pt}{l}p{20pt}{l}p{20pt}{l}p{20pt}{l}p{20pt}{l}}
\toprule
\multirow{2}*{}&\multicolumn{6}{c}{Branin($d_e=2$)}\\
\cline{2-7}
~&\multicolumn{3}{c}{$D$=200}&\multicolumn{3}{c}{$D$=20000}\\
\cline{1-7}
$d$&$2$&$10$&$20$&$2$&$10$&$20$\\
\midrule
REMBO&\textbf{5e\textsuperscript{-7}$\pm$1.3e\textsuperscript{-3}}&0.43$\pm$0.56&0.35$\pm$0.45&\textbf{5e\textsuperscript{-7}$\pm$1.5e\textsuperscript{-7}}&0.77$\pm$0.69&0.80$\pm$1.10\\
SIR-BO&0.11$\pm$ 0.12&0.20$\pm$0.22&0.24$\pm$0.19&0.18$\pm$0.07&\textbf{0.08$\pm$0.05}&0.16$\pm$0.08\\
KISIR-BO&0.25$\pm$0.23&\textbf{0.14$\pm$0.13}&\textbf{0.17$\pm$0.17}&0.14$\pm$0.09&0.13$\pm$0.10&\textbf{0.15$\pm$0.09}\\
\hline
~&\multicolumn{6}{c}{Trimodel($d_e=2$)}\\
\hline
REMBO&\textbf{0.21$\pm$0.66}&3.70$\pm$2.90&10.0$\pm$5.40&0.20$\pm$0.57&4.20$\pm$1.90&5.90$\pm$4.20\\
SIR-BO&\textbf{0.22$\pm$0.15}&0.20$\pm$0.23&0.27$\pm$0.20&\textbf{0.13$\pm$0.12}&0.14$\pm$0.11&\textbf{0.12$\pm$0.14}\\
KISIR-BO&0.30$\pm$0.23&\textbf{0.23$\pm$0.22}&\textbf{0.18$\pm$0.19}&0.16$\pm$0.12&\textbf{0.04$\pm$0.03}&0.14$\pm$0.11\\
\bottomrule
\end{tabular}
\caption{Simple regrets under 500 evaluation budget for $d_e$=2 synthetic functions with different assumed dimension $d$.}
\label{table1}
\end{table}

\begin{figure*}
  \begin{minipage}{5.5cm}
      \includegraphics[width=5.5cm,height=3.5cm]{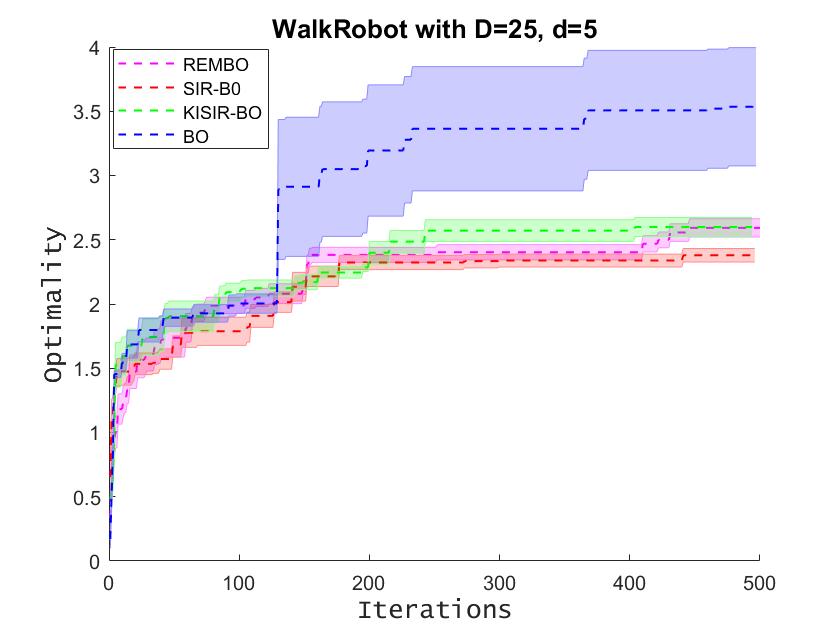}
  \end{minipage}
  \begin{minipage}{5.5cm}
      \includegraphics[width=5.5cm,height=3.5cm]{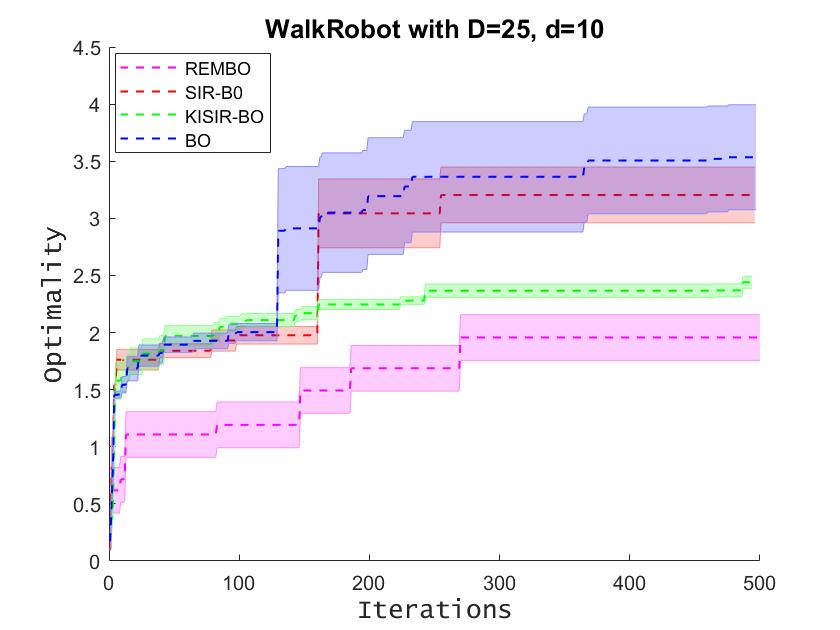}
  \end{minipage} 
  \centering
  \begin{minipage}{5.5cm}
      \includegraphics[width=5.5cm,height=3.5cm]{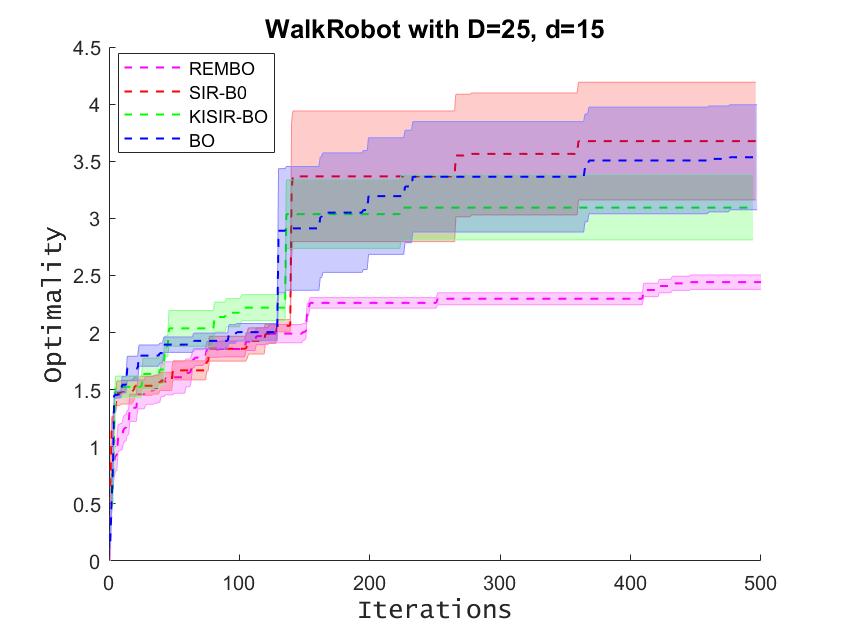}
  \end{minipage}
  \caption{Performance for Walk Robot controlling with different assumed dimensions of subspace. We plot means and $1/4$ standard deviations for each algorithms with 20 independent runs, compared with standard BO optimizing full 25 parameters.}
 \label{figure3}
\end{figure*}

We study the cases with different input dimensions and reduction dimensions: $\{(D=20,d=5),(D=200,d=10),(D=2000,d=10),(D=20000,d=10)\}$. One thing needed be noticed is that, ADD-GP-UCB is not efficient in extremely-high dimensional Bayesian optimization for the combinatorial explosion to find the optimal decomposition ~\cite{baptista2018bayesian} and we do not apply ADD-GP-UCB to cases with $D\geq200$. Figure \ref{figure1} demonstrates the results where our approach get 6 best performances in all 8 cases, and outperforms other baselines in all high dimensions ($D\geq200$) cases. More specifically, ADD-BO performs best in Branin function with median dimension ($D$=20) and REMBO also performs excellent in this case and Branin with remaining dimensions. While they both perform poorly in the nonlinear Trimodel function which is not surprised due to the nonlinear function does not fit the prior linear embedding assumption of REMBO and additive assumption of ADD-BO. Standard BO obtains the best performance in low dimension ($D$=20) for Trimodel function, while it degrades among high dimensions. More interesting, we could find that several basline HDBO methods (REMBO, ADD-BO and Dropout) all perform worse than standard BO in Trimodel function in all dimensions, which again shows that those HDBO methods usually perform poorly when the prior assumption does not fit the objective functions. In the contrast,our SIR-BO and KISIR-BO perform well both in linear function and nonlinear function without the need of prior assumption of objective function, since it could automatically learn the intrinsic sub-structure of objective function during the optimization.

Although REMBO reaches surprised excellent performance in several situations, it is highly dependent on the assumed reduced dimension $d$. To investigate robustness and the effects of the assumed reduced dimension, we further conduct experiments to compare our SIR-BO and KSIR-BO with REMBO under different assumed reduced dimension $d$. Table \ref{table1} analyses the effect of the number of assumed effective dimension $d$, and we can find that REMBO is much dependent on the assumed dimension, where it performs poorly when the assumed dimension $d$ differ from the intrinsic dimension $d_e$ while perform excellently when the assumed dimension equals to intrinsic dimension $d$=$d_e$=$2$. In contrast, our approach is much more robust, where exists no dramatic degrades during the variation of $d$.

\subsection{Optimization on real applications}

Following ~\cite{wang2017batched}, we consider two real-word tasks: training a neural network for Boston dataset with 4 parameters, and controlling a three-link walk robot with 25 parameters ~\cite{westervelt2007feedback}. To emulate the practical fact that we may not know how many parameters the problem dependents on, we augment the number of parameter to \{($D$=20,$d$=5),($D$=200,$d$=10),($D$=2000,$d$=10)\} for Boston dataset and \{($D$=50,$d$=30),($D$=200,$d$=30),($D$=2000,$d$=30)\} for walk robot with dummy variables. Figure \ref{figure2} illustrates the superiority of our SIR-BO and KISIR-BO which could find an effective low-dimensionality for high-dimension hyperparameter configurations. REMBO and ADD-BO both perform poorly in the two real-world applications when the prior assumptions are not in line with the objective functions, while SIR-BO and KISIR-BO perform well since it could learn the intrinsic structure during the optimization.

More interesting, ~\cite{hutter2014efficient} has demonstrated that hyperparameter optimization problem may lie in a low dimensionality of dependent parameters that some hyperparameters are truly unimportant while some hyperparameters are much important. We further conduct experiments on the robot control problem with 25 dependent parameters and optimize it on the subspace of the intrinsic space that $d<d_e$. We study this case with \{($D$=$d_e$=25,$d$=5),($D$=$d_e$=25,$d$=10),($D$=$d_e$=25,$d$=15)\}, and the results in Figure \ref{figure3} demonstrate that the hyperparameter optimization could be conducted in a subspace of dependent parameters (when $d$=15, SIR-BO outperform normal BO with optimizing all dependent parameters), which is in line with the hypothesis of ~\cite{hutter2014efficient}, and suggests that perform a dimension reduction is necessary for hyperparameter optimization problem in machine learning.

\section{Conclusion and future work}

This paper proposed a novel model for high dimensional Bayesian optimization which introduces SIR to automatically learn the intrinsic structure of objective function. In particular, two tricks, reduced SVD and kernelizing input, are developed to reduce computational  complexity and learn non-linear subsets. We theoretically analyze our approach and derive regret bounds. Experimental results demonstrate that our approaches not only perform excellently on high dimension problems, but also are effective on extremely-high dimension problems with $D$=20000. However, our approach still works on high dimensional space to optimize acquisition functions, which is not efficient. In the future works, we try to find more efficient ways to optimize acquisition functions, and also focus on extending our approaches to hyperparameter tuning and network architecture search in Deep Meta-Learning.



\begin{thebibliography}{}

\bibitem[\protect\citeauthoryear{Baptista and
  Poloczek}{2018}]{baptista2018bayesian}
Ricardo Baptista and Matthias Poloczek.
\newblock Bayesian optimization of combinatorial structures.
\newblock In {\em International Conference on Machine Learning}, pages
  462--471, 2018.

\bibitem[\protect\citeauthoryear{Bergstra \bgroup \em et al.\egroup
  }{2011}]{bergstra2011algorithms}
James~S Bergstra, R{\'e}mi Bardenet, Yoshua Bengio, and Bal{\'a}zs K{\'e}gl.
\newblock Algorithms for hyper-parameter optimization.
\newblock In {\em Advances in neural information processing systems}, pages
  2546--2554, 2011.

\bibitem[\protect\citeauthoryear{Chen \bgroup \em et al.\egroup
  }{2012}]{chen2012joint}
Bo~Chen, Rui~M. Castro, and Andreas Krause.
\newblock Joint optimization and variable selection of high-dimensional
  gaussian processes.
\newblock pages 1423--1430, 2012.

\bibitem[\protect\citeauthoryear{Cunningham and
  Ghahramani}{2015}]{cunningham2015linear}
John~P Cunningham and Zoubin Ghahramani.
\newblock Linear dimensionality reduction: Survey, insights, and
  generalizations.
\newblock {\em The Journal of Machine Learning Research}, 16(1):2859--2900,
  2015.

\bibitem[\protect\citeauthoryear{Deisenroth \bgroup \em et al.\egroup
  }{2015}]{deisenroth2015gaussian}
Marc~Peter Deisenroth, Dieter Fox, and Carl~Edward Rasmussen.
\newblock Gaussian processes for data-efficient learning in robotics and
  control.
\newblock {\em IEEE transactions on pattern analysis and machine intelligence},
  37(2):408--423, 2015.

\bibitem[\protect\citeauthoryear{Djolonga \bgroup \em et al.\egroup
  }{2013}]{djolonga2013high}
Josip Djolonga, Andreas Krause, and Volkan Cevher.
\newblock High-dimensional gaussian process bandits.
\newblock In {\em Advances in Neural Information Processing Systems}, pages
  1025--1033, 2013.

\bibitem[\protect\citeauthoryear{Hutter \bgroup \em et al.\egroup
  }{2014}]{hutter2014efficient}
Frank Hutter, Holger Hoos, and Kevin Leyton-Brown.
\newblock An efficient approach for assessing hyperparameter importance.
\newblock In {\em International Conference on Machine Learning}, pages
  754--762, 2014.

\bibitem[\protect\citeauthoryear{Kandasamy \bgroup \em et al.\egroup
  }{2015}]{kandasamy2015high}
Kirthevasan Kandasamy, Jeff Schneider, and Barnab{\'a}s P{\'o}czos.
\newblock High dimensional bayesian optimisation and bandits via additive
  models.
\newblock In {\em International Conference on Machine Learning}, pages
  295--304, 2015.

\bibitem[\protect\citeauthoryear{Kandasamy \bgroup \em et al.\egroup
  }{2018}]{kandasamy2018neural}
Kirthevasan Kandasamy, Willie Neiswanger, Jeff Schneider, Barnabas Poczos, and
  Eric~P Xing.
\newblock Neural architecture search with bayesian optimisation and optimal
  transport.
\newblock In {\em Advances in Neural Information Processing Systems}, pages
  2020--2029, 2018.

\bibitem[\protect\citeauthoryear{Li \bgroup \em et al.\egroup
  }{2016}]{li2016high}
Chun-Liang Li, Kirthevasan Kandasamy, Barnab{\'a}s P{\'o}czos, and Jeff
  Schneider.
\newblock High dimensional bayesian optimization via restricted projection
  pursuit models.
\newblock In {\em Artificial Intelligence and Statistics}, pages 884--892,
  2016.

\bibitem[\protect\citeauthoryear{Li \bgroup \em et al.\egroup
  }{2017}]{li2017high}
Cheng Li, Sunil Gupta, Santu Rana, Vu~Nguyen, Svetha Venkatesh, and Alistair
  Shilton.
\newblock High dimensional bayesian optimization using dropout.
\newblock In {\em Proceedings of the 26th International Joint Conference on
  Artificial Intelligence}, pages 2096--2102. AAAI Press, 2017.

\bibitem[\protect\citeauthoryear{Li}{1991}]{li1991sliced}
Ker-Chau Li.
\newblock Sliced inverse regression for dimension reduction.
\newblock {\em Journal of the American Statistical Association},
  86(414):316--327, 1991.

\bibitem[\protect\citeauthoryear{Lizotte}{2008}]{lizotte2008practical}
Daniel~James Lizotte.
\newblock {\em Practical bayesian optimization}.
\newblock PhD thesis, University of Alberta, 2008.

\bibitem[\protect\citeauthoryear{Rasmussen and
  Williams}{2004}]{rasmussen-et-al:gp}
Carl~Edward Rasmussen and Christopher K.~I. Williams.
\newblock {\em Gaussian Processes for Machine Learning}.
\newblock MIT Press, Cambridge, Massachusetts, 2004.

\bibitem[\protect\citeauthoryear{Rolland \bgroup \em et al.\egroup
  }{2018}]{rolland2018high}
Paul Rolland, Jonathan Scarlett, Ilija Bogunovic, and Volkan Cevher.
\newblock High-dimensional bayesian optimization via additive models with
  overlapping groups.
\newblock In {\em International Conference on Artificial Intelligence and
  Statistics}, pages 298--307, 2018.

\bibitem[\protect\citeauthoryear{Snoek \bgroup \em et al.\egroup
  }{2012}]{snoek2012practical}
Jasper Snoek, Hugo Larochelle, and Ryan~P Adams.
\newblock Practical bayesian optimization of machine learning algorithms.
\newblock In {\em Advances in neural information processing systems}, pages
  2951--2959, 2012.

\bibitem[\protect\citeauthoryear{Srinivas \bgroup \em et al.\egroup
  }{2012}]{srinivas2012information}
Niranjan Srinivas, Andreas Krause, Sham~M Kakade, and Matthias~W Seeger.
\newblock Information-theoretic regret bounds for gaussian process optimization
  in the bandit setting.
\newblock {\em IEEE Transactions on Information Theory}, 58(5):3250--3265,
  2012.

\bibitem[\protect\citeauthoryear{Tan and Mukherjee}{2018}]{tan2018subspace}
Zilong Tan and Sayan Mukherjee.
\newblock Subspace-induced gaussian processes.
\newblock {\em arXiv preprint arXiv:1802.07528}, 2018.

\bibitem[\protect\citeauthoryear{Ulmasov \bgroup \em et al.\egroup
  }{2016}]{ulmasov2016bayesian}
Doniyor Ulmasov, Caroline Baroukh, Benoit Chachuat, Marc~Peter Deisenroth, and
  Ruth Misener.
\newblock Bayesian optimization with dimension scheduling: Application to
  biological systems.
\newblock In {\em Computer Aided Chemical Engineering}, volume~38, pages
  1051--1056. Elsevier, 2016.

\bibitem[\protect\citeauthoryear{Wang \bgroup \em et al.\egroup
  }{2013}]{wang2013bayesian}
Ziyu Wang, Masrour Zoghi, Frank Hutter, David Matheson, Nando De~Freitas,
  et~al.
\newblock Bayesian optimization in high dimensions via random embeddings.
\newblock In {\em IJCAI}, pages 1778--1784, 2013.

\bibitem[\protect\citeauthoryear{Wang \bgroup \em et al.\egroup
  }{2017}]{wang2017batched}
Zi~Wang, Chengtao Li, Stefanie Jegelka, and Pushmeet Kohli.
\newblock Batched high-dimensional bayesian optimization via structural kernel
  learning.
\newblock In {\em International Conference on Machine Learning}, pages
  3656--3664, 2017.

\bibitem[\protect\citeauthoryear{Westervelt \bgroup \em et al.\egroup
  }{2007}]{westervelt2007feedback}
Eric~R Westervelt, Christine Chevallereau, Jun~Ho Choi, Benjamin Morris, and
  Jessy~W Grizzle.
\newblock {\em Feedback control of dynamic bipedal robot locomotion}.
\newblock CRC press, 2007.

\bibitem[\protect\citeauthoryear{Wu \bgroup \em et al.\egroup
  }{2007}]{wu2007regularized}
Qiang Wu, F~Liang, and S~Mukherjee.
\newblock Regularized sliced inverse regression for kernel models.
\newblock Technical report, Technical Report 07-25, ISDS, Duke Univ, 2007.

\bibitem[\protect\citeauthoryear{Wu}{2008}]{wu2008kernel}
Han-Ming Wu.
\newblock Kernel sliced inverse regression with applications to classification.
\newblock {\em Journal of Computational and Graphical Statistics},
  17(3):590--610, 2008.

\bibitem[\protect\citeauthoryear{Yeh \bgroup \em et al.\egroup
  }{2009}]{yeh2009nonlinear}
Yi-Ren Yeh, Su-Yun Huang, and Yuh-Jye Lee.
\newblock Nonlinear dimension reduction with kernel sliced inverse regression.
\newblock {\em IEEE Transactions on Knowledge and Data Engineering},
  21(11):1590--1603, 2009.

\end{thebibliography}
\end{document}